%% file: icml2023.tex
\documentclass{article}

\usepackage{microtype}
\usepackage{graphicx}
\usepackage{subfigure}
\usepackage{booktabs} 
\usepackage{threeparttable}

\usepackage{hyperref}

\usepackage[accepted]{icml2023}

\usepackage{amsmath}
\usepackage{amssymb}
\usepackage{mathtools}
\usepackage{amsthm}
\usepackage{microtype}

\usepackage[capitalize,noabbrev]{cleveref}

\theoremstyle{plain}

\theoremstyle{definition}

\theoremstyle{remark}

\usepackage[disable,textsize=tiny]{todonotes}

\usepackage{url}
\usepackage{wrapfig}
\usepackage{multirow}
\usepackage{makecell}
\usepackage{colortbl}
\usepackage{tablefootnote}
\usepackage{diagbox}
\input{math_commands.tex}

\input{abbr_commands}

\icmltitlerunning{A Closer Look at Self-Supervised Lightweight Vision Transformers}

\begin{document}

\twocolumn[
\icmltitle{A Closer Look at Self-Supervised Lightweight Vision Transformers}



\icmlsetsymbol{equal}{*}

\begin{icmlauthorlist}
\icmlauthor{Shaoru Wang}{casia,ucas}
\icmlauthor{Jin Gao}{casia,ucas}
\icmlauthor{Zeming Li}{megvii}
\icmlauthor{Xiaoqin Zhang}{wenzhou}
\icmlauthor{Weiming Hu}{casia,ucas,shanghaitech}
\end{icmlauthorlist}

\icmlaffiliation{casia}{State Key Laboratory of Multimodal Artificial Intelligence Systems, Institute of Automation, Chinese Academy of Sciences}
\icmlaffiliation{ucas}{School of Artificial Intelligence, University of Chinese Academy of Sciences}
\icmlaffiliation{megvii}{Megvii Research}
\icmlaffiliation{wenzhou}{Key Laboratory of Intelligent Informatics for Safety \& Emergency of Zhejiang Province, Wenzhou University}
\icmlaffiliation{shanghaitech}{School of Information Science and Technology, ShanghaiTech University}

\icmlcorrespondingauthor{Jin Gao}{jin.gao@nlpr.ia.ac.cn}

\icmlkeywords{Self-Supervised Learning, Vision Transformer, Lightweight Network, Knowledge Distillation}

\vskip 0.3in
]



\printAffiliationsAndNotice{}  

\begin{abstract}
Self-supervised learning on large-scale Vision Transformers (ViTs) as pre-training methods has achieved promising downstream performance. Yet, how much these pre-training paradigms promote lightweight ViTs' performance is considerably less studied. In this work, we develop and benchmark several self-supervised pre-training methods on image classification tasks and some downstream dense prediction tasks. We surprisingly find that if proper pre-training is adopted, even vanilla lightweight ViTs show comparable performance to previous SOTA networks with delicate architecture design. It breaks the recently popular conception that vanilla ViTs are not suitable for vision tasks in lightweight regimes. We also point out some defects of such pre-training, \eg, failing to benefit from large-scale pre-training data and showing inferior performance on data-insufficient downstream tasks. Furthermore, we analyze and clearly show the effect of such pre-training by analyzing the properties of the layer representation and attention maps for related models. Finally, based on the above analyses, a distillation strategy during pre-training is developed, which leads to further downstream performance improvement for MAE-based pre-training. 
Code is available at \hyperlink{https://github.com/wangsr126/mae-lite}{https://github.com/wangsr126/mae-lite}.
\end{abstract}

\vspace{-10pt}
\section{Introduction}
Self-supervised learning (SSL) has shown great progress in representation learning without heavy reliance on expensive labeled data. SSL focuses on various pretext tasks for pre-training. Among them, several works \citep{mocov1, mocov2, BYOL, swav, mocov3, dino} based on contrastive learning (CL) have achieved comparable or even better accuracy than supervised pre-training when transferring the learned representations to downstream tasks. Recently, another trend focuses on masked image modeling (MIM) \citep{beit, mae, ibot}, which perfectly fits Vision Transformers (ViTs) \citep{vit} for vision tasks, and achieves improved generalization performance. Most of these works, however, involve large networks with little attention paid to smaller ones. Some works \citep{SEED, CompRess, oss} focus on CL on small convolutional networks (ConvNets) and improve the performance by distillation. However, the pre-training of lightweight ViTs is considerably less studied. 

Efficient neural networks are essential for modern on-device computer vision. Recent studies on achieving top-performing lightweight models mainly focus on designing network architectures \citep{mobilenetv2, mobilenetv3, levit, xcit, pit, cait, mobilevit, mobileformer, edgevit}, while little attention is paid to how to optimize the training strategies for these models. We believe the latter is also of vital importance, and the utilization of pre-training is one of the most hopeful approaches along this way, since it has achieved great progress on large models. To this end, we develop and benchmark recently popular self-supervised pre-training methods, \eg, CL-based MoCo-v3 \citep{mocov3} and MIM-based MAE \citep{mae}, along with fully-supervised pre-training for lightweight ViTs as the baselines on ImageNet and other classification tasks, as well as some dense prediction tasks, \eg, object detection and segmentation. We surprisingly find that \emph{if proper pre-training is adopted, even vanilla lightweight ViTs show comparable performance to previous SOTA networks with delicate design}, \eg, we achieve 79.0\% top-1 accuracy on ImageNet with vanilla ViT-Tiny (5.7M). The finding is intriguing since the result indicates that proper pre-training could bridge the performance gap between naive network architectures and delicately designed ones to a great extent, while naive architectures usually have faster inference speed, by getting rid of some complicated operators. We also point out some defects of such pre-training, \eg, \emph{failing to benefit from large-scale pre-training data} and \emph{showing inferior performance on data-insufficient downstream tasks}. 

These findings motivate us to dive deep into the working mechanism of these pre-training methods for lightweight ViTs. More specifically, we introduce a variety of model analysis methods to study the pattern of layer behaviors during pre-training and fine-tuning, and investigate what really matters for downstream performance. First, we find that \emph{lower layers of the pre-trained models matter more than higher ones if sufficient downstream data is provided, while higher layers matter in data-insufficient downstream tasks}. Second, we observe that \emph{the pre-training with MAE makes the attention of the downstream models more local and concentrated, \ie, introduces locality inductive bias, which may be the key to the performance gain}. 
Based on the above analyses, we also develop a distillation strategy for MAE-based pre-training, which significantly improves the pre-training of lightweight ViTs. Better downstream performance is achieved especially on data-insufficient classification tasks and detection tasks.

\section{Preliminaries and Experimental Setup}

\paragraph{ViTs.} We use ViT-Tiny \citep{deit} 
to examine the effect of the pre-training on downstream performance, which contains 5.7M parameters. We adopt the vanilla architecture, consisting of a patch embedding layer and 12 Transformer blocks with an embedding dimension of 192, except that the number of heads is increased to 12 as we find it can improve the model's expressive power. 
ViT-Tiny is chosen for study because it is an ideal experimental object, on which almost all existing pre-training methods can be directly applied. And it has a rather naive architecture: non-hierarchical, and with low human inductive bias in design. Thus the influence of the model architecture design on our analyses can be eliminated to a great extent.

\paragraph{Evaluation Metrics.}
We adopt \emph{fine-tuning} as the default evaluation protocol considering that it is highly correlated with utility \cite{newell2020useful}, in which all the layers are tuned by initializing them with the pre-trained models. 
By default, we do the evaluation on ImageNet \citep{ImageNet} by fine-tuning on the training set and evaluating on the validation set. Several other downstream classification datasets (\eg, Flowers \citep{flower}, Aircraft \citep{aircraft}, CIFAR100 \citep{cifar}, \etc) and object detection and segmentation tasks on COCO \citep{mscoco} are also exploited for comparison. For a more thorough study, analyses based on \emph{linear probing} evaluation are presented in \cref{sec:appdix-lp}.

\paragraph{Compared Methods.}

\emph{Baseline}: We supervisedly train a ViT-Tiny from scratch for 300 epochs on the training set of ImageNet-1k (dubbed IN1K). 
It achieves 74.5\% top-1 accuracy on the validation set of ImageNet-1k, surpassing that in the original architecture (72.2\% \citep{deit}) through modifying the number of heads to 12 from 3, and further reaches 75.8\% by adopting our improved training recipe (see \cref{sec:appdix-eval}), which finally serves as our strong baseline to examine the pre-training. We denote this model from supervised training as DeiT-Tiny.

\emph{MAE}: MAE \citep{mae} is selected as a representative for MIM-based pre-training methods, which has a simple framework with low training cost. We largely follow the design of MAE except that the encoder is altered to ViT-Tiny. Several basic factors and components are adjusted to fit the smaller encoder (see \cref{sec:appdix-mae}). 
By default, we do pre-training on IN1K for 400 epochs, and denote the pre-trained model as MAE-Tiny.

\emph{MoCov3}: We also implement a contrastive SSL pre-training counterpart, MoCo-v3 \citep{mocov3}, which is selected for its simplicity. We also do 400-epoch pre-training and denote the pre-trained model as MoCov3-Tiny.
Details are provided in \cref{sec:appdix-mocov3}.

Some other methods, \eg, MIM-based SimMIM \cite{simmim} and CL-based DINO \cite{dino} are also involved, but are moved to \cref{sec:appdix-more} due to space limitation.

\section{How Well Does Pre-Training Work on Lightweight ViTs?}\label{sec:exp}
In this section, we first benchmark the aforementioned pre-trained models on ImageNet, and then further evaluate their transferability to other datasets and tasks.

\subsection{Benchmarks on ImageNet Classification Tasks}
\begin{table*}[t]
    \centering
    \input{tables/pretrain}
    \vspace{-13pt}
\end{table*}

\paragraph{Which pre-training method performs best?}
We first develop and benchmark the pre-training methods on ImageNet, involving the baseline that does not adopt any pre-training, supervised pre-training on the training set of ImageNet-21k (a bigger and more diverse dataset, as roughly ten times the size of IN1K, dubbed IN21K) and the aforementioned self-supervised pre-training with MoCo-v3 and MAE.
As reported in \cref{tbl:pretrain}, most of these supervised and self-supervised pre-training methods improve the downstream performance, whilst \emph{MAE outperforms others and consumes moderate training cost}. 
The results indicate that the vanilla ViTs have great potential, which can be unleashed via proper pre-training. It encourages us to further explore how the enhanced ViTs perform compared to recent SOTA ConvNets and ViT derivatives.

\paragraph{How do the enhanced ViTs with pre-training rank among SOTA lightweight networks?}

To answer the question, we further compare the enhanced ViT-Tiny with MAE pre-training to previous lightweight ConvNets and ViT derivatives. 
We report top-1 accuracy along with the model parameter count and the throughput in \cref{tbl:sota}. We denote the fine-tuned model based on MAE-Tiny as MAE-Tiny-FT.
Specifically, we extend the fine-tuning epochs to 1000 following \citet{deit} and adopt relative position embedding. Under this strong fine-tuning recipe, the pre-training still contributes a 1.2 performance gain, ultimately reaching 79.0\% top-1 accuracy. It sets a new record for lightweight vanilla ViTs, even without distillation during the supervised training phase on IN1K.
It can also be seen that the pre-training can accelerate the downstream convergence, which helps to surpass that trained from scratch for 1000 epochs (77.8\%) with only 300-epoch fine-tuning (78.5\%).

We conclude that \emph{the enhanced ViT-Tiny is on par with or even outperforms most previous ConvNets and ViT derivatives with comparable parameters or throughput}. 
This demonstrates that we can also achieve SOTA performance based on a naive network architecture by adopting proper pre-training, 
rather than designing complex ones. Significantly, naive architecture usually has faster inference speed and is friendly to deployment.

We also notice that there are some works applying supervised pre-training \citep{ridnik2021imagenet21k}, CL-based self-supervised pre-training \citep{SEED} and MIM-based self-supervised pre-training \citep{convnextv2} on lightweight ConvNets. However, we find that ViT-Tiny benefits more from the pre-training (\eg, +1.2 vs. +0.5 for ConvNeXt V2-F). We attribute it to that the plain architecture of ViT-Tiny with less artificial design may possess more model capacity.

\paragraph{Can the pre-training benefit from more data?}
\begin{table}[t]
    \input{tables/pretrain-dataset}
    \vspace{-15pt}
\end{table}

One may be curious about whether it is possible to achieve better downstream performance by involving more pre-training data, as it does on large models. Unfortunately, the answer is no for the examined pre-training methods. We consider IN21K, a much larger dataset. The number of pre-training iterations is kept constant for a fair comparison. However, few improvements are observed for both MoCo-v3 and MAE as shown in \cref{tbl:pretrain-dataset}. We further consider two subsets of IN1K containing 1\% and 10\% of the total examples (1\% IN1K and 10\% IN1K) balanced in terms of classes \citep{assran2021semi} and one subset with long-tailed class distribution \citep{imagenet-lt} (IN1K-LT). Surprisingly, marginal performance declines are observed for MAE when pre-training on these subsets, showing more robustness than MoCo-v3 in terms of the pre-training data scale and class distribution.

\begin{table*}[t]
\input{tables/sota}
\vspace{-11pt}
\end{table*}
\begin{table*}[h!]
\input{tables/transfer}
\vspace{-11pt}
\end{table*}

\subsection{Benchmarks on Transfer Performance}\label{sec:transfer}
We further examine the transferability of these models pre-trained on IN1K, involving their transfer performance on some other classification tasks and dense prediction tasks. In addition to the self-supervised MAE-Tiny and MoCov3-Tiny, DeiT-Tiny is also involved, as a fully-supervised counterpart which is trained on IN1K for 300 epochs.

\paragraph{Can the pre-trained models transfer well on data-insufficient tasks?}

We introduce several classification tasks \citep{flower, oxford-pet, aircraft, stanford-cars, cifar, inat} to investigate their transferability. We conduct the transfer evaluation by fine-tuning these pre-trained models on these datasets (see \cref{sec:appdix-transfer} for more details). 
 As shown in \cref{tbl:transfer}, 
 using various pre-training methods shows better performance than using random initialization, but the relative superiority and inferiority comparisons between these pre-training methods exhibit distinct characteristics from those on ImageNet.
 We find that \emph{downstream data scale matters}. The self-supervised pre-training approaches achieve downstream performance far behind the fully-supervised counterpart, while the performance gap is narrowed more or less as the data scale of the downstream task increases. Moreover, MAE even shows inferior results to MoCo-v3. We conjecture that it is due to their different layer behaviors during pre-training and fine-tuning, which will be discussed in detail in the following section.

\paragraph{Can the pre-trained models transfer well on dense prediction tasks?}
For a more thorough study, we further conduct evaluations on downstream object detection and segmentation tasks on COCO \citep{mscoco}, based on \citet{li2021benchmarking} (see \cref{sec:appdix-det} for details) with different pre-trained models as initialization of the backbone. The results are shown in \cref{tbl:transfer}. The self-supervised pre-training also lags behind the fully-supervised counterpart.

\begin{figure*}[t]
    \begin{minipage}[t][][b]{0.7\textwidth}
    \centering
    \includegraphics[width=1.0\textwidth]{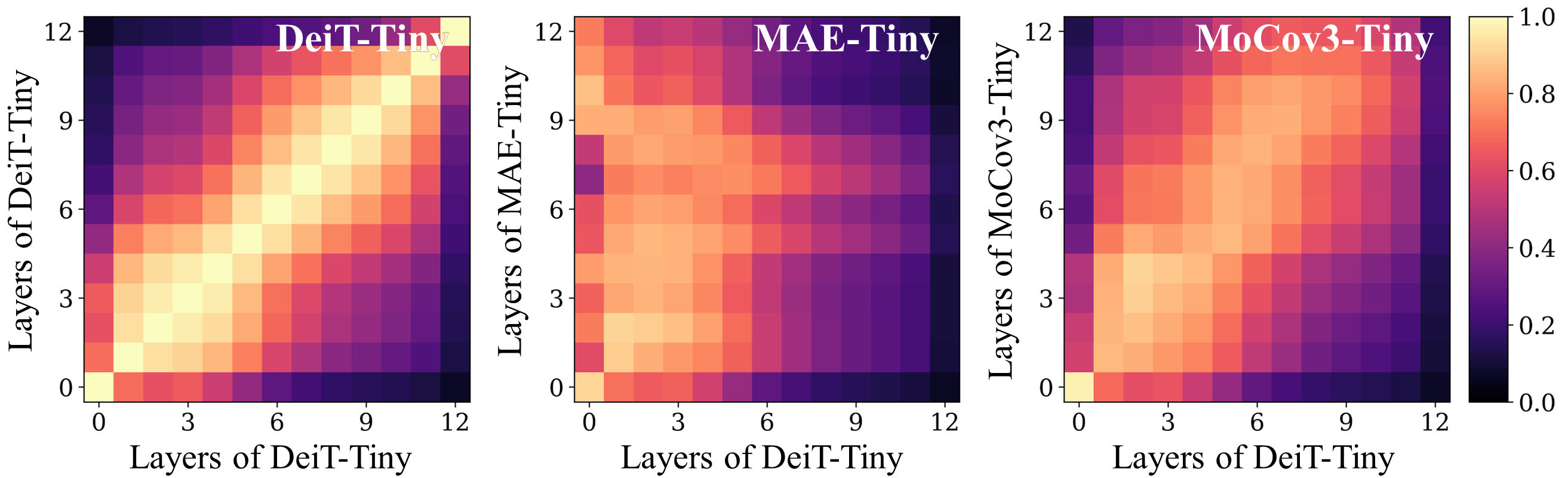}
    \vspace{-19pt}
    \caption{\small{\textbf{Layer representation similarity} within and across models as heatmaps, with x and y axes indexing the layers (the 0 index indicates the patch embedding layer), and higher values indicate higher similarity.
    }}
    \label{fig:cmp}
    \end{minipage}\hspace{2mm}
    \begin{minipage}[t][][b]{0.28\textwidth}
    \centering
    \includegraphics[width=1.0\textwidth]{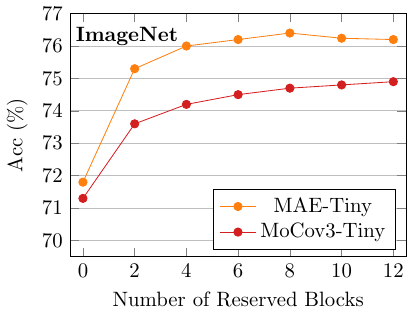}
    \vspace{-18pt}
    \caption{\small{Lower layers of pre-trained models contribute to most gains on downstream ImageNet dataset.
    }}
    \label{fig:drop}
    \end{minipage}
    \vspace{-10pt}
\end{figure*}

\section{Revealing the Secrets of the Pre-Training}\label{sec:work}

In this section, we introduce some model analysis methods to study the pattern of layer behaviors during pre-training and fine-tuning, and investigate what matters for downstream performances.

\subsection{Layer Representation Analyses}\label{sec:rep-analyses}
We first adopt Centered Kernel Alignment (CKA) method\footnote{\hyperlink{https://github.com/AntixK/PyTorch-Model-Compare}{https://github.com/AntixK/PyTorch-Model-Compare}} \citep{cortes2012algorithms, nguyen2020wide} to analyze the layer representation similarity across and within networks. Specifically, CKA computes the normalized similarity in terms of the Hilbert-Schmidt Independence Criterion (HSIC \citep{song2012feature}) between two feature maps or representations, which is invariant to the orthogonal transformation of representations and isotropic scaling (detailed in \cref{sec:appdix-analysis-methods}). 



\paragraph{Lower layers matter more than higher ones if sufficient downstream data is provided. }

We visualize the layer representation similarity between several pre-trained models and DeiT-Tiny as heatmaps in \cref{fig:cmp}. We choose DeiT-Tiny, a classification model fully-supervisedly trained on IN1K, as the reference because we consider the higher similarity of the examined model's layer to that of DeiT-Tiny indicates its more relevance to recognition. Although the similarity does not directly indicate whether the downstream performance is good, it indeed reflects the pattern of layer representation to a certain extent. The similarity within DeiT-Tiny is also presented (the left column).

First, We observe a relatively high similarity between MAE-Tiny and DeiT-Tiny for lower layers, while low similarity for higher layers. In \cref{sec:appdix-ref}, we observe similar phenomena with several additional supervisedly trained ViTs as the reference models.
It indicates that fewer semantics are extracted for MAE-Tiny at a more abstract level in higher layers. In contrast, MoCov3-Tiny aligns DeiT-Tiny well across almost all layers. However, the fine-tuning evaluation in \cref{tbl:pretrain} shows that adopting the MAE-Tiny as initialization improves the performance more significantly than MoCov3-Tiny. Thus, we hypothesize that \emph{lower layers matter much more than higher ones for the pre-trained models}. In order to verify the hypothesis, we design another experiment by only reserving several leading blocks of pre-trained models and randomly initializing the others, and then fine-tuning them on IN1K (for the sake of simplicity, we only fine-tune these models for 100 epochs). \cref{fig:drop} shows that reserving only a certain number of leading blocks achieves a significant performance gain over randomly initializing all the blocks (\ie, totally training from scratch) for both MAE-Tiny and MoCov3-Tiny. Whereas, further reserving higher layers leads to marginal gain for MAE-Tiny and MoCov3-Tiny, which demonstrates our hypothesis.

\begin{figure*}[t]
    \centering
    \includegraphics[width=0.29\textwidth]{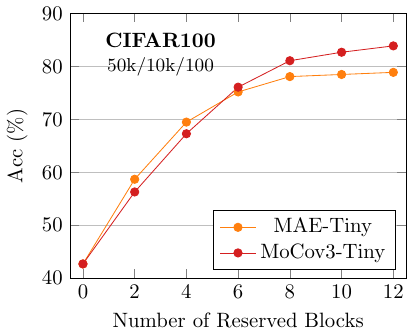}\hspace{2mm}
    \includegraphics[width=0.29\textwidth]{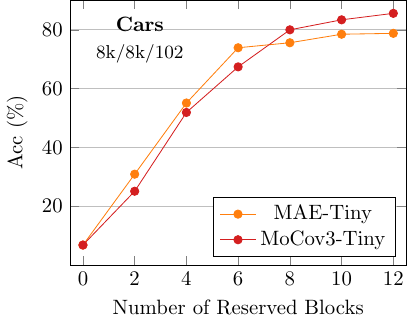}\hspace{2mm}
    \includegraphics[width=0.29\textwidth]{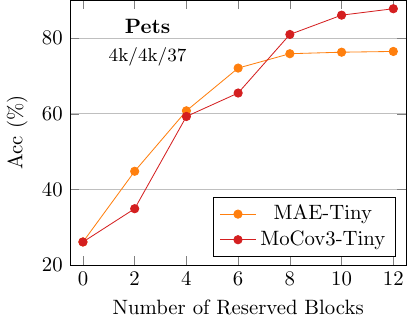}
    \vspace{-10pt}
    \caption{\small{The contributions on performance gain from higher layers of pre-trained models increase as the downstream dataset scale shrinks, which indicates that higher layers matter in data-insufficient downstream tasks.
    }}
    \label{fig:drop-transfer}
\vspace{-10pt}
\end{figure*}

\paragraph{Higher layers matter in data-insufficient downstream tasks.}
Previous works \citep{deit, raghu2021vision} demonstrate the importance of a relatively large dataset scale for fully-supervised high-performance ViTs with large model sizes. We also observe a similar phenomenon on lightweight ViTs even when the self-supervised pre-training is adopted as discussed in \cref{sec:transfer}. It motivates us to study the key factor of downstream performance on data-insufficient tasks.

We conduct similar experiments as those in \cref{fig:drop} on small-scale downstream datasets. The results are shown in \cref{fig:drop-transfer}. We observe consistent performance improvement as the number of reserved pre-trained models' blocks increases. And the smaller the dataset scale, the more the performance benefits from the higher layers. It demonstrates that higher layers are still valuable and matter in data-insufficient downstream tasks. Furthermore, we observe comparable performance for the transfer performance of MAE-Tiny and MoCov3-Tiny when only a certain number of lower layers are reserved, while MoCov3-Tiny surpasses when higher layers are further reserved. It indicates that the higher layers of MoCov3-Tiny work better than MAE-Tiny on data-insufficient downstream tasks, which is also consistent with our CKA-based analyses shown in \cref{fig:cmp}, that MoCov3-Tiny learns more semantics at an abstract level relevant to recognition in higher layers (high similarity to reference recognition models in higher layers) than MAE-Tiny. 

\begin{figure*}[t]
    \begin{minipage}[t][][b]{0.664\textwidth}
    \centering
    \includegraphics[width=1.0\textwidth]{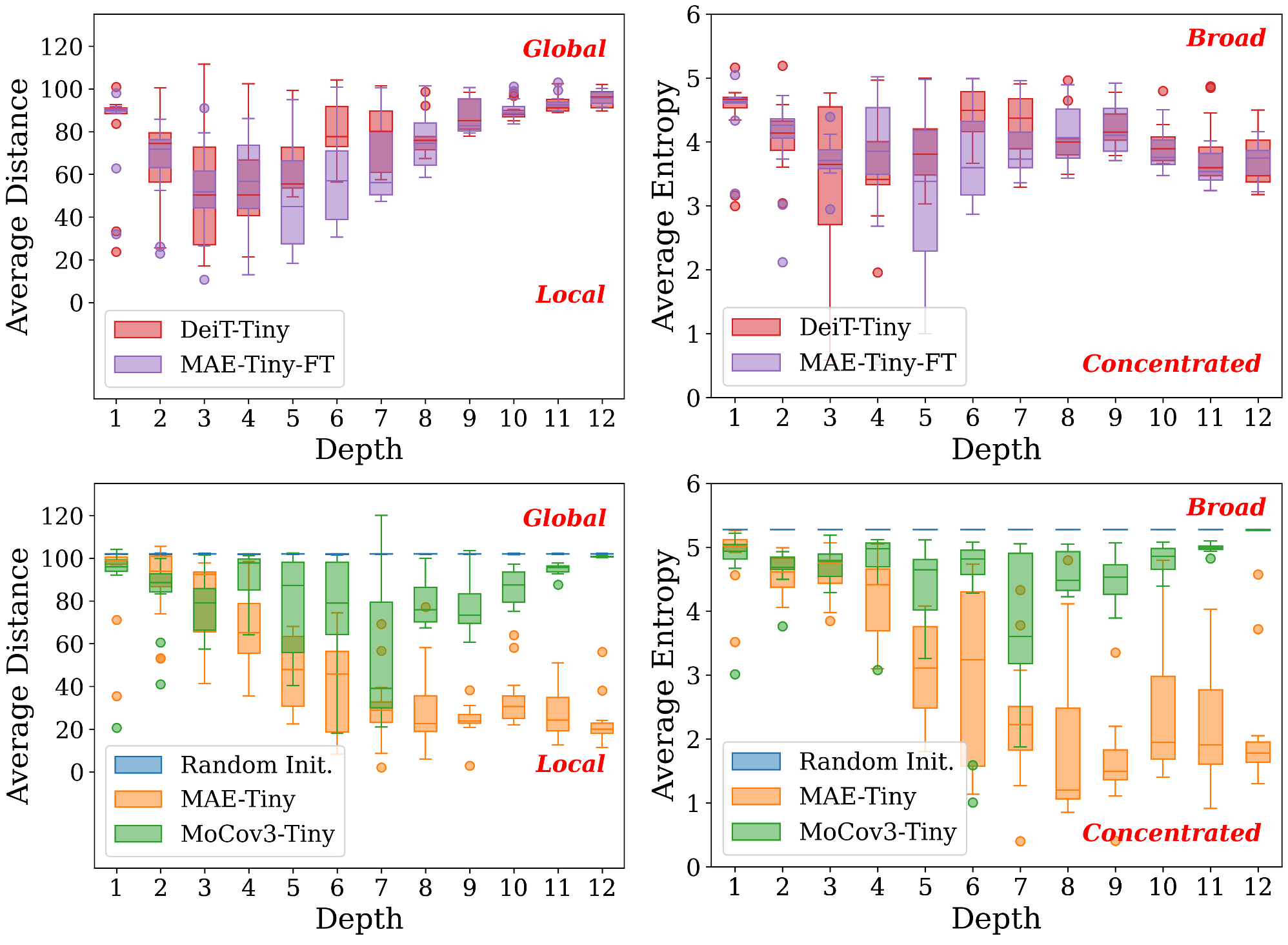}
    \vspace{-21.7pt}
    \caption{\small{\textbf{Attention distance and entropy analyses}. We visualize the distributions of the average attention distance and entropy across all tokens in different attention heads \wrt the layer number with box-whisker plots.}}
    \label{fig:attn}
    \end{minipage}\hspace{2mm}
    \begin{minipage}[t][][b]{0.303\textwidth}
    \centering
        \vspace{-2pt}
        \includegraphics[width=1.0\textwidth]{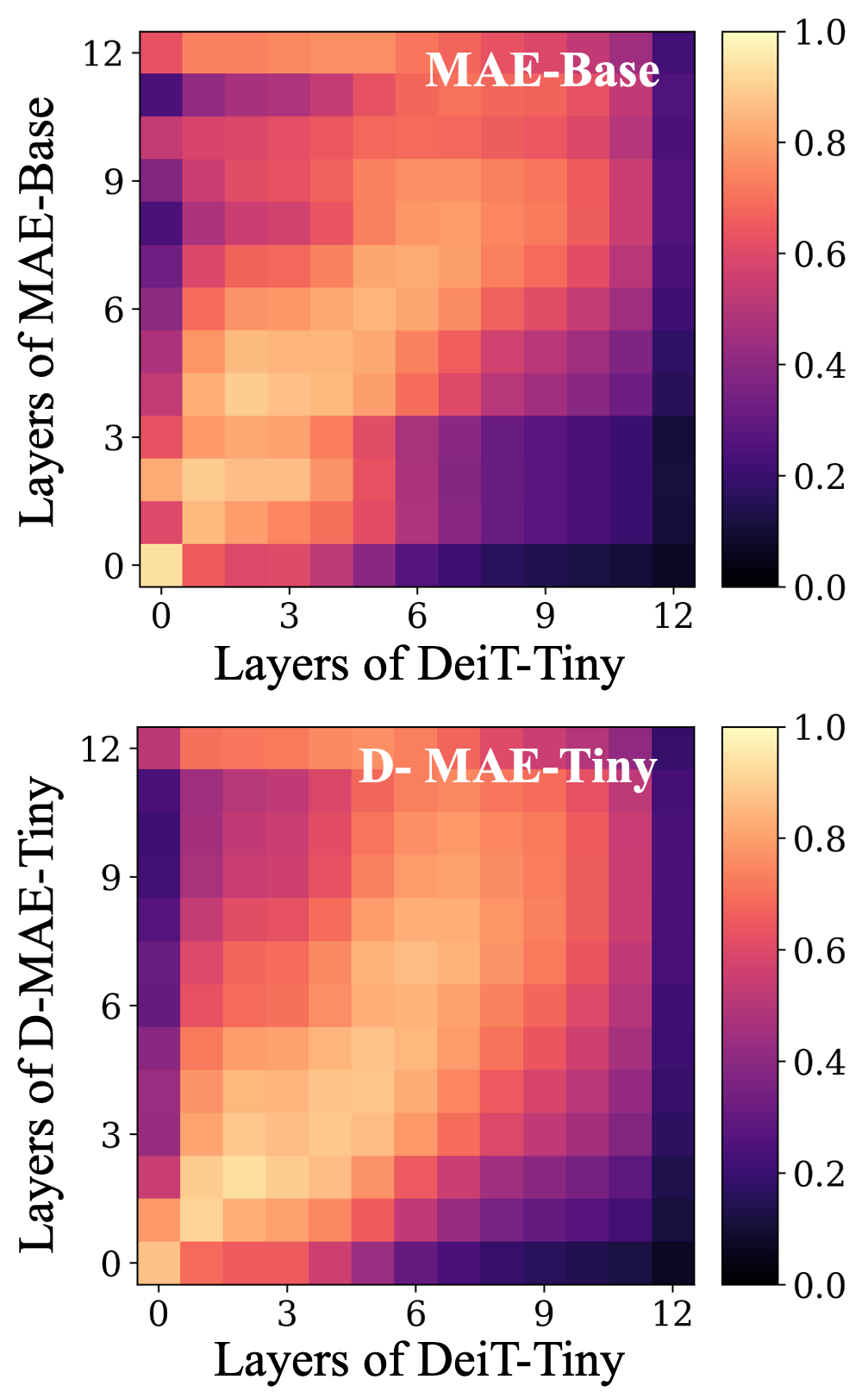}
        \vspace{-19pt}
        \caption{\small{Distillation compresses the good representation of the teacher (MAE-Base) to the student (D-MAE-Tiny).
        }}
        \label{fig:cmp-distill}
    \end{minipage}
\vspace{-10pt}
\end{figure*}

\subsection{Attention Map Analyses}
The attention maps reveal the behaviors for aggregating information in the attention mechanism of ViTs, which are computed from the compatibility of queries and keys by dot-product operation. We introduce two metrics for further analyses on the pre-trained models, \ie, \emph{attention distance} and \emph{attention entropy}. 
The attention distance for the $j$-th token of $h$-th head is calculated as:
\begin{align}
    \mD_{h,j}=\sum_i \softmax(\mA_h)_{i,j}\mG_{i,j},
\end{align}
where $\mA_h\in\mathbb{R}^{l\times l}$ is the attention map for the $h$-th attention head, and $\mG_{i,j}$ is the Euclidean distance between the spatial locations of the $i$-th and $j$-th tokens. $l$ is the number of tokens. And the attention entropy is calculated as:
\begin{align}
    \mE_{h,j}=-\sum_i \softmax(\mA_h)_{i,j}\mathrm{log}(\softmax(\mA_h)_{i,j}),
\end{align}

Specifically, the attention distance reveals how much local \vs global information is aggregated, and a lower distance indicates that each token focuses more on neighbor tokens. The attention entropy reveals the concentration of the attention distribution, and lower entropy indicates that each token attends to fewer tokens.  We analyze the distributions of the average attention distance and entropy across all the tokens in different attention heads, as shown in \cref{fig:attn}. 

\begin{figure}[t]
    \centering
    \includegraphics[width=0.34\textwidth]{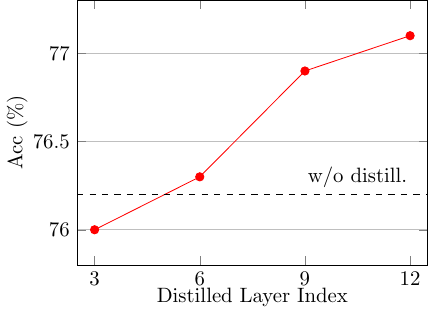}
    \vspace{-10pt}
    \caption{\small{Distillation on attention maps of higher layers improves performance most.} }
    \label{fig:distill} 
\vspace{-15pt}
\end{figure}

\paragraph{The pre-training with MAE makes the attention of the downstream models more local and concentrated.} 
First, we compare MAE-Tiny-FT with DeiT-Tiny. The former adopts MAE-Tiny as initialization and then is fine-tuned on IN1K, and the latter is supervisedly trained from scratch (Random Init.) on IN1K. As shown in \cref{fig:attn}, we observe very similar attention behaviors between them, except that the attention of MAE-Tiny-FT (the \textcolor[RGB]{160,120,196}{purple} box-whisker) is more local (with lower attention distance) and concentrated (with lower attention entropy) in middle layers compared with DeiT-Tiny (the \textcolor[RGB]{219,64,65}{red} box-whisker). We attribute it to the introduction of the MAE-Tiny as pre-training (the \textcolor[RGB]{255,165,85}{orange} box-whisker), which has lower attention distance and entropy, and may bring locality inductive bias compared with random initialization (the \textcolor[RGB]{57,135,189}{blue} box-whisker). It is noteworthy that the locality inductive bias does not mean that tokens in all attention heads attend to solely a few nearby tokens. The attention distance and entropy for different heads are still distributed in a wide range (except for several last layers), which indicates that the heads have diverse specializations, making the models aggregate both local and global tokens with both concentrated and broad focuses.

Then, we focus on the comparison between MAE-Tiny and MoCov3-Tiny, trying to give some explanations for their diverse downstream performances observed in \cref{sec:exp}. As shown in \cref{fig:attn}, we observe that MoCov3-Tiny (the \textcolor[RGB]{68,171,68}{green} box-whisker) generally has more global and broad attention than MAE-Tiny (the \textcolor[RGB]{255,165,85}{orange} box-whisker). Even several leading blocks have a narrower range of attention distance and entropy than MAE-Tiny. We think this characteristic of MoCov3-Tiny makes the downstream fine-tuning with it as initialization take ``shortcuts", \ie, directly paying attention to global features and overlooking local patterns, which may be unfavorable for fine-grained recognition. It leads to inferior downstream performance on ImageNet, but fair on Flowers, CIFAR100, \etc, for which the ``shortcuts" may be barely adequate.
As for MAE-Tiny, its distinct behaviors in higher layers with rather low attention distance and entropy may make it hard to transfer to data-insufficient downstream tasks, thus resulting in inferior performance on these tasks.


\section{Distillation Improves Pre-Trained Models}\label{sec:distill}

In the previous section, we have conjectured that it is hard for MAE to learn good representation relevant to recognition in higher layers, which results in unsatisfactory performance on data-insufficient downstream tasks.
A natural question is that can it gain more semantic information by scaling up the models. We further examine a large pre-trained model, MAE-Base \citep{mae}, and find it achieves a better alignment with the reference model, as shown in the top subfigure of \cref{fig:cmp-distill}. It indicates that \emph{it is possible to extract features relevant to recognition in higher layers for the scaled-up encoder in MAE pre-training.}
These observations motivate us to compress the knowledge of large pre-trained models to tiny ones with knowledge distillation under the MIM framework.

\paragraph{Distillation methods.}

\begin{table*}[t]
\input{tables/transfer-distill}
\vspace{-15pt}
\end{table*}

Specifically, a pre-trained MAE-Base \citep{mae} is introduced as the teacher network. The distillation loss is constructed based on the similarity between the attention maps of the corresponding teacher's and student's layers. It is formulated as:
\begin{align}
    L_{\mathrm{attn}} &= \mathrm{MSE}(\bm{A}^T, \bm{M}\bm{A}^S),
\end{align}
where $\bm{A}^T\in \mathbb{R}^{h\times l \times l}$ and $\bm{A}^S\in \mathbb{R}^{h'\times l \times l}$ refer to the attention maps of the corresponding teacher's and student's layers, with $h$ and $h'$ attention heads respectively. $l$ is the number of tokens. A learnable mapping matrix $\bm{M}\in \mathbb{R}^{h\times h'}$ is introduced to align the number of heads. $\mathrm{MSE}$ denotes mean squared error.

During the pre-training, the teacher processes the same unmasked image patches as the student encoder. The parameters of the student network are updated based on the joint backward gradients from the distillation loss and the original MAE's reconstruction loss, while the teacher's parameters remain frozen throughout the pre-training process.


\paragraph{Distill on lower or higher layers?}
We first examine applying the above layer-wise distillation on which pair of teacher's and student's layers contributes to the most performance gain. We conduct experiments by constructing the above attention-based distillation loss between pair of layers at $1/4$, $2/4$, $3/4$, or $4/4$ depth of the teacher and student respectively, \ie, the 3rd, 6th, 9th, or 12th layer for both the teacher (MAE-Base) and the student (MAE-Tiny).
As shown in \cref{fig:distill}, distilling on the attention maps of the last transformer blocks promotes the performance most, surpassing those distilling on lower layers (for the sake of simplicity, we only fine-tune the pre-trained models on IN1K for 100 epochs). It is consistent with the analyses in \cref{sec:work}. Specifically, the lower layers learn good representation themselves during the pre-training with MAE, and thus distilling on these layers contributes to marginal improvement, while the higher layers rely on a good teacher to guide them to capture rich semantic features.

\paragraph{Distillation improves downstream performance.} We further evaluate the distilled pre-trained model on several downstream tasks. For simplicity, we only apply distillation on the last layers. The resulting model is denoted as D-MAE-Tiny. The visualization result at the bottom of \cref{fig:cmp-distill} shows that the good representation relevant to the recognition of the teacher is compressed to the student. Especially the quality of higher layers is improved. The distillation contributes to better downstream performance as shown in \cref{tbl:transfer-distill}, especially on data-insufficient classification tasks and dense prediction tasks.
In \cref{sec:appdix-distill}, we also show that our distillation technique can help other ViT students beyond ViT-Tiny to achieve better downstream performance.

\section{Related Works}
\paragraph{Self-supervised learning (SSL)} focuses on different pretext tasks \citep{rotation_pred, image_colorization, jigsaw, instance_discrimination} for pre-training without using manually labeled data. Among them, contrastive learning (CL) has been popular and shows promising results on various convolutional networks (ConvNets) \citep{mocov1, mocov2, BYOL, swav} and ViTs \citep{mocov3, dino}.
Recently, methods based on masked image modeling (MIM) achieve state-of-the-art on ViTs \citep{mae, beit, ibot}.
It has been demonstrated that these methods can scale up well on larger models, while their performance on lightweight ViTs is seldom investigated. 

\paragraph{Vision Transformers (ViTs)} \citep{vit} apply a Transformer architecture (a stack of attention modules \citep{attention}) on image patches and show very competitive results in various visual tasks \citep{deit,swin,li2022exploring}. The performance of ViTs has been largely improved thanks to better training recipes \citep{deit, steiner2021train, touvron2022deit}. As for lightweight ViTs, most works focus on integrating ViTs and ConvNets \citep{levit, pit, mobilevit, mobileformer}, while few works focus on how to optimize the networks. 

\paragraph{Knowledge Distillation} is a mainstream approach for model compression \citep{compression}, in which a large teacher network is trained first and then a more compact student network is optimized to approximate the teacher 
\citep{hinton2015distilling, fitnets}. \citet{deit} achieves better accuracy on ViTs by adopting a ConvNet as the teacher. With regard to the compression of the pre-trained networks, some works \citep{distillbert, tinybert, minilmv2, mobilebert} attend to distill large-scale pre-trained language models. In the area of computer vision, a series of works \citep{SEED, CompRess, oss} focus on transferring knowledge of large pre-trained networks based on CL to lightweight ConvNets. There are few works focusing on improving the quality of lightweight pre-trained ViTs based on MIM by distillation thus far.

\section{Discussions}\label{sec:discussion}
\paragraph{Limitations.}
Our study is restricted to classification tasks and some dense-prediction tasks. We leave the exploration of more tasks for further work.
\paragraph{Conclusions.}
We investigate the self-supervised pre-training of lightweight ViTs, and demonstrate the usefulness of the advanced lightweight ViT pre-training strategy in improving the performance of downstream tasks, even comparable to most delicately-designed SOTA networks on ImageNet. 
Some properties about the pre-training are revealed, \eg, these methods fail to benefit from large-scale pre-training data, and show more dependency on the downstream dataset scale. We also present some insights on what matters for downstream performance.
They may indicate potential future directions in improving pre-training on lightweight models, the value of which has also been demonstrated as it guides the design of our proposed distillation strategy and helps to achieve much better downstream performance. We expect our research may provide useful experience and advance the study of self-supervised learning on lightweight ViTs.

\textbf{Acknowledgment.} 
The authors would like to thank the anonymous reviewers for their valuable comments and suggestions.
This work was supported in part by the National Key R\&D Program of China (Grant No. 2020AAA0105802, 2020AAA0105800), the Natural Science Foundation of China (Grant No. U22B2056, 61972394, U2033210, 62036011, 62192782, 61721004, 62172413), the Beijing Natural Science Foundation (Grant No. L223003, JQ22014), the Major Projects of Guangdong Education Department for Foundation Research and Applied Research (Grant No. 2017KZDXM081, 2018KZDXM066), the Guangdong Provincial University Innovation Team Project (Grant No. 2020KCXTD045), the Zhejiang Provincial Natural Science Foundation (Grant No. LDT23F02024F02). Jin Gao was also supported in part by the Youth Innovation Promotion Association, CAS.

\bibliography{icml2023}
\bibliographystyle{icml2023}

\input{appendix}


\end{document}

%% file: math_commands.tex
\usepackage{amsmath,amsfonts,bm}

\def\eqref#1{equation~\ref{#1}}

\def\1{\bm{1}}

\def\vs{{\bm{s}}}

\def\mA{{\bm{A}}}

\def\mD{{\bm{D}}}
\def\mE{{\bm{E}}}

\def\mG{{\bm{G}}}

\DeclareMathAlphabet{\mathsfit}{\encodingdefault}{\sfdefault}{m}{sl}
\SetMathAlphabet{\mathsfit}{bold}{\encodingdefault}{\sfdefault}{bx}{n}

\newcommand{\softmax}{\mathrm{softmax}}



%% file: tables/pretrain.tex
\setlength{\tabcolsep}{14pt}
\begin{center}
\renewcommand{\arraystretch}{1.0} 
{
\caption{\small{\textbf{Comparisons on pre-training methods}. 
We report top-1 accuracy on the validation set of ImageNet-1k. 
IN1K and IN21K indicate the training set of ImageNet-1k and ImageNet-21k. The pre-training time is measured on 8$\times$V100 GPU machine. 
`ori.' represents the supervised training recipe from \citet{deit} and `impr.' represents our improved recipe (see \cref{sec:appdix-eval}).
}
}
\label{tbl:pretrain}
\vspace{-10pt}
\resizebox{1.0\textwidth}{!}{
\begin{threeparttable}
\small
\begin{tabular}{cccc|cc}
\toprule
\multicolumn{4}{c|}{\textbf{\ \quad\quad Pre-training}} & \multicolumn{2}{c}{\textbf{\textbf{Fine-tuning}}}\\
\textbf{Methods} & \textbf{Data} & \textbf{Epochs} & \textbf{Time (hour)} & \textbf{recipe} & \textbf{Top-1 Acc. (\%)} \\
\midrule
- & - & - & - & ori. & 74.5 \\
- & - & - & - & impr. & 75.8 \\
Supervised \citep{steiner2021train} & IN21K w/ labels & 30 & 20 & impr. & 76.9 \\
Supervised \citep{steiner2021train} & IN21K w/ labels & 300 & 200 & impr. & 77.8 \\
MoCo-v3 \citep{mocov3} & IN1K w/o labels & 400 & 52 & impr. & $\enspace$76.8\dag \\
MAE \citep{mae} & IN1K w/o labels & 400 & 23 & impr. & \textbf{78.0} \\
\bottomrule
\end{tabular}
\begin{tablenotes}
    \footnotesize
    \item[\dag] \textls[-30]{Global average pooling is used instead of the default configuration based on the class token during the fine-tuning. See \cref{sec:appdix-eval} for details.}
    \end{tablenotes}
\end{threeparttable}
}
}
\end{center}

%% file: tables/pretrain-dataset.tex
\setlength{\tabcolsep}{4pt}
\begin{center}
\renewcommand{\arraystretch}{0.8} 
\caption{\small{\textbf{Effect of pre-training data}. Top-1 accuracy is reported. 
}}
\label{tbl:pretrain-dataset}
\small
\begin{tabular}{c|c|c}
\toprule
\textbf{Datasets} & \textbf{MoCo-v3} & \textbf{MAE} \\
\midrule
IN1K & 76.8 & 78.0 \\
1\% IN1K & 76.2 \small\colorgreen{(-0.6)} & 77.9 \small\colorgreen{(-0.1)} \\
10\% IN1K & 76.5 \small\colorgreen{(-0.3)} & 78.0 \small\colorred{(+0.0)}\\
IN1K-LT & 76.1 \small\colorgreen{(-0.7)} & 77.9 \small\colorgreen{(-0.1)}\\
IN21K & 76.9 \small\colorred{(+0.1)} & 78.0 \small\colorred{(+0.0)}\\
\bottomrule
\end{tabular}
\end{center}

%% file: tables/sota.tex
\setlength{\tabcolsep}{3pt}
\begin{center}
\renewcommand{\arraystretch}{0.98} 
{
\caption{\small{\textbf{Comparisons with previous SOTA networks on ImageNet-1k. }
We report top-1 accuracy
along with throughput and parameter count. The throughput is borrowed from timm \citep{rw2019timm}, which is measured on a single RTX 3090 GPU with a batch size fixed to 1024 and mixed precision. `\dag' indicates that distillation is adopted during the supervised training (or fine-tuning). `$^\star$' indicates the original architecture of \mbox{ViT-Tiny} (the number of attention heads is 3). 
}
}
\label{tbl:sota}
\small
\begin{tabular}{c|c|c|c|c|c}
\toprule
 \multirow{2}{*}{\textbf{Methods}} & \multirow{2}{*}{\textbf{pre-train data}} & \textbf{fine-tuning} & \multirow{2}{*}{\textbf{\#param.}} & \textbf{throughput} & \textbf{Accuracy}\\
  & & \textbf{epochs} & & (image/s) & Top-1 (\%) \\
\midrule
\multicolumn{5}{c}{\emph{ConvNets}} \\
\midrule
ResNet-18 \citep{resnet} & - & 100 & 11.7M & 8951 & 69.7\\
ResNet-50 \citep{resnet,wightman2021resnet} & - & 600 & 25.6M & 2696 & 80.4\\
\midrule
EfficientNet-B0 \citep{efficientnet} & - & 450 & 5.3M & 5369 & 77.7\\
EfficientNet-B0 \citep{SEED} & IN1K w/o labels & 450 & 5.3M & 5369 & 77.2 \tiny\colorgreen{(-0.5)}\\
EfficientNet-B1 \citep{efficientnet} & - & 450 & 7.8M & 2953 & 78.8\\
\midrule
MobileNet-v2 \citep{mobilenetv2} & - & 480 & 3.5M & 7909 & 72.0\\
MobileNet-v3 \citep{mobilenetv3} & - & 600 & 5.5M & 9113 & 75.2\\
MobileNet-v3\dag \citep{ridnik2021imagenet21k} & IN21K & 600 & 5.5M & 9113 & 78.0\\
\midrule
ConvNeXt V1-F \citep{convnext} & - & 600 & 5.2M & - & 77.5\\
ConvNeXt V2-F \citep{convnextv2} & - & 600 & 5.2M & 1816 & 78.0\\
ConvNeXt V2-F \citep{convnextv2} & IN1K w/o labels & 600 & 5.2M & 1816 & 78.5 \tiny\colorred{(+0.5)}\\
\midrule
\multicolumn{5}{c}{\emph{Vision Transformers Derivative}} \\
\midrule
LeViT-128 \citep{levit} & - & 1000 & 9.2M & 13276 & 78.6\\
LeViT-192 \citep{levit} & - & 1000 & 11.0M & 11389 & 80.0\\
\midrule
XCiT-T12/16\dag \citep{xcit} & - & 400 & 6.7M & 3157 & 78.6 \\
\midrule
PiT-Ti\dag \citep{pit} & - & 1000 & 5.1M & 4547 & 76.4\\
\midrule
CaiT-XXS-24\dag \citep{cait} & - & 400 & 12.0M & 1351 & 78.4\\
\midrule
Swin-1G \citep{swin, mobileformer} & - & 450 & 7.3M & - & 77.3\\
\midrule
Mobile-Former-294M \citep{mobileformer} & - & 450 & 11.4M & - & 77.9\\
\midrule
MobileViT-S \citep{mobilevit} & - & 300 & 5.6M & 1900 & 78.3\\
\midrule
EdgeViT-XS \citep{edgevit} & - & 300 & 6.7M & - & 77.5\\
\midrule
\multicolumn{5}{c}{\emph{Vanilla Vision Transformers}} \\
\midrule
DeiT-Tiny$^\star$ \citep{deit} & - & 300 & 5.7M & 4844 & 72.2\\
DeiT-Tiny$^\star$\dag \citep{deit} & - & 1000 & 5.7M & 4764 & 76.6\\
\bottomrule
DeiT-Tiny & - & 300 & 5.7M & 4020 & 76.2\\
\rowcolor{lightgray}
MAE-Tiny-FT & IN1K w/o labels & 300 & 5.7M & 4020 & 78.5 \tiny\colorred{(+2.3)}\\
DeiT-Tiny & - & 1000 & 5.7M & 4020 & 77.8 \\
\rowcolor{lightgray}
MAE-Tiny-FT & IN1K w/o labels & 1000 & 5.7M & 4020 & 79.0 \tiny\colorred{(+1.2)}\\
\toprule
\end{tabular}}
\end{center}

%% file: tables/transfer.tex
\setlength{\tabcolsep}{4pt}
\begin{center}
\renewcommand{\arraystretch}{0.9} 
{
\caption{\small{\textbf{Transfer evaluation on classification tasks and dense-prediction tasks}. Self-supervised pre-training approaches generally show inferior performance to the fully-supervised counterpart. Top-1 accuracy is reported for classification tasks and AP is reported for object detection (det.) and instance segmentation (seg.) tasks.
The description of each dataset is represented as (train-size/test-size/\#classes).}}
\label{tbl:transfer}
\small
\begin{tabular}{c|cccccccc}
\toprule
\multirow{2}{*}{\diagbox{\textbf{Init.}}{\textbf{Datasets}}} & \textbf{Flowers} & \textbf{Pets} & \textbf{Aircraft} & \textbf{Cars} & \textbf{CIFAR100} & \textbf{iNat18} & \textbf{COCO}\small{(det.)} & \textbf{COCO}\small{(seg.)}\\
& \scriptsize{(2k/6k/102)} & \scriptsize{(4k/4k/37)} & \scriptsize{(7k/3k/100)} & \scriptsize{(8k/8k/196)} & \scriptsize{(50k/10k/100)} & \scriptsize{(438k/24k/8142)} & \multicolumn{2}{c}{\scriptsize{(118k/50k/80)}}\\
 \midrule
Random & 30.2 & 26.1 & 9.4 & 6.8 & 42.7 & 58.7 & 32.7 & 28.9\\
\midrule
\colorgray{\textit{supervised}} \\
DeiT-Tiny & \textbf{96.4} & \textbf{93.1} & 73.5 & \textbf{85.6} & \textbf{85.8} & \textbf{63.6} & \textbf{40.4} & \textbf{35.5} \\
\midrule
\colorgray{\textit{self-supervised}} \\
MoCov3-Tiny & 94.8 & 87.8 & \textbf{73.7} & 83.9 & 83.9 & 54.5 & 39.7 & 35.1 \\
MAE-Tiny & 85.8 & 76.5 & 64.6 & 78.8 & 78.9 & 60.6 & 39.9 & 35.4 \\
\bottomrule
\end{tabular}}
\end{center}

%% file: tables/transfer-distill.tex
\setlength{\tabcolsep}{3pt}
\begin{center}
\renewcommand{\arraystretch}{0.8} 
{
\caption{\small{\textbf{Distillation improves downstream performance} on classification tasks and object detection and segmentation tasks. Top-1 accuracy is reported for classification tasks and AP is reported for object detection (det.) and instance segmentation (seg.) tasks.
}}
\label{tbl:transfer-distill}
\small
\begin{tabular}{c|ccccccccc}
\toprule
\diagbox{\small\textbf{Init.}}{\small\textbf{Datasets}} & \textbf{Flowers} & \textbf{Pets} & \textbf{Aircraft} & \textbf{Cars} & \textbf{CIFAR100} & \textbf{iNat18} & \textbf{ImageNet} & \textbf{COCO}\small{(det.)} & \textbf{COCO}\small{(seg.)}\\
 \midrule
\colorgray{\textit{supervised}} \\
DeiT-Tiny & \textbf{96.4} & \textbf{93.1} & 73.5 & 85.6 & \textbf{85.8} & \textbf{63.6} & - & 40.4 & 35.5 \\
\midrule
\colorgray{\textit{self-supervised}} \\
MAE-Tiny & 85.8 & 76.5 & 64.6 & 78.8 & 78.9 & 60.6 & 78.0 & 39.9 & 35.4 \\
\rowcolor{lightgray}
D-MAE-Tiny & 95.2 \tiny\colorred{(+9.4)} & 89.1 \tiny\colorred{(+12.6)} & \textbf{79.2} \tiny\colorred{(+14.6)} & \textbf{87.5} \tiny\colorred{(+8.7)} & 85.0 \tiny\colorred{(+6.1)} & \textbf{63.6} \tiny\colorred{(+3.0)} & \textbf{78.4} \tiny\colorred{(+0.4)} & \textbf{42.3} \tiny\colorred{(+2.4)} & \textbf{37.4} \tiny\colorred{(+2.0)}\\
\toprule
\end{tabular}}
\end{center}

%% file: appendix.tex
\clearpage
\appendix
\onecolumn
\setcounter{table}{0}
\renewcommand{\thetable}{A\arabic{table}}
\setcounter{equation}{0}
\renewcommand{\theequation}{A\arabic{equation}}
\setcounter{figure}{0}
\renewcommand{\thefigure}{A\arabic{figure}}
\pagenumbering{Roman}


\section{Experimental Details}\label{sec:appdix-detail}
\subsection{Evaluation Details for MAE and MoCo-v3 on ImageNet}\label{sec:appdix-eval}
We follow the common practice of supervised ViT training \citep{deit} for fine-tuning evaluation except for some hyper-parameters of augmentation. The default setting is in \cref{tbl:appdix-eval-detail}. We use the linear \emph{lr} scaling rule \citep{imagenet_in_1_hour}: \emph{lr} = base \emph{lr}$\times$batchsize / 256. We use layer-wise \emph{lr} decay following \citep{beit, mae}, and the decay rate is tuned respectively for MAE and MoCo-v3. 

Besides, we use global average pooling (GAP) after the final block during the fine-tuning of both the MAE and MoCo-v3-based pre-trained models, which is, however, not the common practice for MoCo-v3 \citep{mocov3}. We adopt it as it significantly helps to surpass the model using the original configuration based on a class token (76.8\% \vs 73.7\% top-1 accuracy) for the lightweight ViT-Tiny.


\begin{table}[ht]
\centering
\vspace{-10pt}
    \begin{minipage}[t][][b]{0.62\textwidth}
    \input{appdix/eval-detail}
    \end{minipage}\hspace{1mm}
    \begin{minipage}[t][][b]{0.36\textwidth}
    \input{appdix/mocov3}
    \end{minipage}
\vspace{-15pt}
\end{table}

\subsection{Pre-Training Details of MAE}\label{sec:appdix-mae}
Our experimental setup on MAE largely follows those of MAE \citep{mae}, including the optimizer, learning rate, batch size, argumentation, \etc. But several basic factors and components are adjusted to fit the smaller encoder. 
We find MAE prefers a much more lightweight decoder when the encoder is small, thus a decoder with only one Transformer block is adopted by default and the width is 192. We sweep over 5 masking ratios \{0.45, 0.55, 0.65, 0.75, 0.85\} and find 0.75 achieves the best performance.


\subsection{Pre-Training Details of MoCo-v3}\label{sec:appdix-mocov3}
We reimplement MoCo-v3 \citep{mocov3} with ViT-Tiny as encoder and largely follow the original setups. The default setting is in \cref{tbl:appdix-mocov3}. 


\citet{mocov3} observes that instability is a major issue that impacts self-supervised ViT training and causes mild degradation in accuracy, and a simple trick by adopting fixed random patch projection (the first layer of a ViT model) is proposed to improve stability in practice. However, we find that stability is not the main issue for small networks. Higher performance is achieved with a learned patch projection layer. Thus, this technique is not used by default. 


\subsection{Transfer Evaluation Details on Classification Tasks}\label{sec:appdix-transfer}
We evaluate several pre-trained models with transfer learning in order to measure the generalization ability of these models. We use 6 popular vision datasets: Flowers-102 (Flowers for short) \citep{flower}, Oxford-\uppercase\expandafter{\romannumeral3}T Pets (Pets) \citep{oxford-pet}, FGVC-Aircraft (Aircraft) \citep{aircraft}, Stanford Cars (Cars) \citep{stanford-cars}, Cifar100 \citep{cifar}, iNaturalist 2018 (iNat18) \citep{inat}. For all these datasets except iNat18, we fine-tune with SGD (momentum=0.9), and the batch size is set to 512. The learning rates are swept over 3 candidates and the training epochs are swept over 2 candidates per dataset as detailed in \cref{tbl:appdix-transfer-detail}. We adopt a cosine decay learning rate schedule \citep{sgdr} with a linear warm-up. we resize images to 224 $\times$ 224. We adopt random resized crop and random horizontal flipping as augmentations and do not use any regularization (\eg, weight decay, dropout, or the stochastic depth regularization technique \citep{huang2016deep}). For iNat18, we follow the same training configurations as those on ImageNet.
\begin{table}[t]
\input{appdix/transfer-detail}
\vspace{-10pt}
\end{table}

\subsection{Transfer Evaluation Details on Dense Prediction Tasks}\label{sec:appdix-det}
We reproduce the setup in \citep{li2021benchmarking}, except for replacing the backbone with ViT-Tiny and decreasing the input image size from 1024 to 640 to make it trainable on a single machine with 8 NVIDIA V100. We fine-tune for up to 100 epochs on COCO \citep{mscoco}, with different pre-trained models as initialization of the backbone. We do not use layer-wise \emph{lr} decay since we find it useless for the tiny backbone on the detection tasks. The weight decay is 0.05 and the stochastic depth regularization \citep{huang2016deep} is not used.

\subsection{Analysis Methods}\label{sec:appdix-analysis-methods}
\begin{wrapfigure}{r}{0.32\textwidth}
    \centering
    \vspace{-12pt}
    \includegraphics[width=0.21\textwidth]{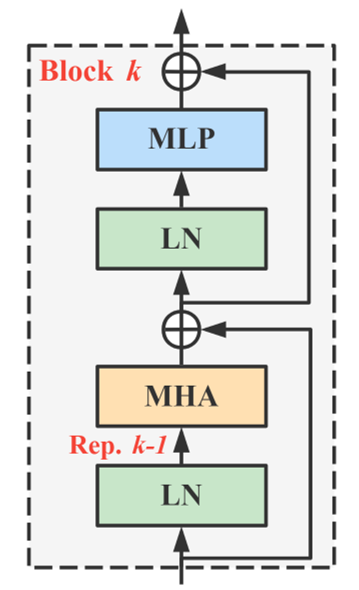}
    \caption{\small{\textbf{Transformer block.}}}
    \label{fig:appdix-transformer}
    \vspace{-10pt}
\end{wrapfigure}
We adopt the Centered Kernel Alignment (CKA) metric to analyze the representation similarity ($S_{rep}$) within and across networks. Specifically, CKA takes two feature maps (or representations) $\bm{X}$ and $\bm{Y}$ as input and computes their normalized similarity in terms of the Hilbert-Schmidt Independence Criterion (HSIC) as 
\begin{align}
    S_{rep}(\bm{X},\bm{Y})=\mathrm{CKA}(\bm{K},\bm{L})=\frac{\mathrm{HSIC}(\bm{K},\bm{L})}{\sqrt{\mathrm{HSIC}(\bm{K},\bm{K})\mathrm{HSIC}(\bm{L},\bm{L})}},
\end{align}
where $\bm{K}=\bm{X}\bm{X}^\mathrm{T}$ and $\bm{L}=\bm{Y}\bm{Y}^\mathrm{T}$ denote the Gram matrices for the two feature maps. A minibatch version is adopted by using an unbiased estimator of HSIC \citep{nguyen2020wide} to work at scale with our networks. We select the normalized version of the output representation of each Transformer block (consisting of a multi-head self-attention (MHA) block and an MLP block). Specifically, we select the feature map after the first LayerNorm (LN) \citep{layer-norm} of the next block as the representation of this Transformer block as depicted in \cref{fig:appdix-transformer}.

\section{More Analyses on the Pre-Training}
\subsection{Analyses with More Models as Reference}\label{sec:appdix-ref}

\begin{figure*}[ht]
    \centering
\includegraphics[width=1.0\textwidth]{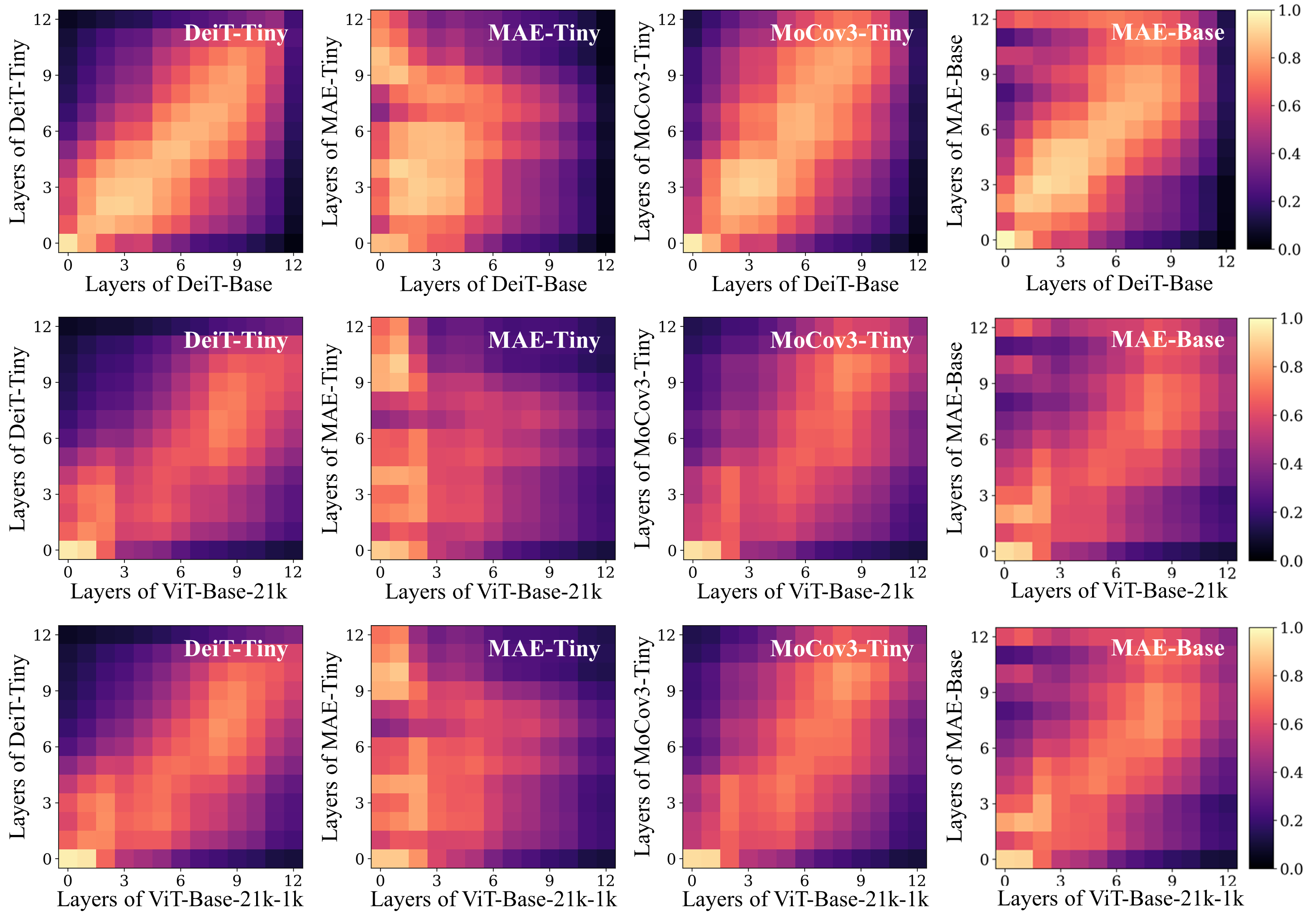}
    \vspace{-18pt}
    \caption{\textbf{Layer representation analyses} with DeiT-Base (supervisedly trained on IN1K, the top row), ViT-Base-21k (supervisedly trained on IN21K, the middle row), and ViT-Base-21k-1k (supervisedly pre-trained on IN21K and fine-tuned on IN1K, the bottom row) as the reference models.}\label{fig:appdix-cmp}
\vspace{-12pt}
\end{figure*}


In \cref{sec:work}, the analyses are mainly conducted by adopting the supervisedly trained DeiT-Tiny as the reference model. Here, we additionally introduce stronger recognition models as references to demonstrate the generalizability of our analyses. Specifically, we use ViT-Base models trained with various recipes as references, \eg, DeiT-Base (supervisedly trained on IN1K following \citet{deit} and achieves 82.0\% top-1 accuracy on ImageNet), ViT-Base-21k (supervisedly trained on IN21K following \citet{steiner2021train}), ViT-Base-21k-1k (first pre-trained on IN21K and then fine-tuned on IN1K following \citet{steiner2021train}, achieving 84.5\% top-1 accuracy on ImageNet). The layer representation similarity is presented in \cref{fig:appdix-cmp}.

First, we observe that our default reference model, DeiT-Tiny, is aligned well with these larger models (as shown in the left column of \cref{fig:appdix-cmp}). We conjecture that the supervisedly trained ViTs generally have similar layer representation structures. Based on these stronger reference models, we observe similar phenomena for MAE-Tiny and MoCov3-Tiny as discussed in \cref{sec:work}, which demonstrates the robustness of our analyses and conclusions \wrt different reference models.

Then, we analyze the larger MAE-Base with these newly introduced models as references, as shown in the last column of \cref{fig:appdix-cmp}. We observe that MAE-Base still aligns relatively well with these much stronger recognition models, which supports our claim in \cref{sec:distill} that \emph{it is possible to extract features relevant to recognition in higher layers for the scaled-up encoder in MAE pre-training}. It is the prerequisite for the improvement of the pre-trained models from the proposed distillation.

\subsection{Analyses Based on Linear Probing Evaluation}\label{sec:appdix-lp}
Our analyses are mainly based on the \emph{fine-tuning} evaluation. In this section, we present some experimental results based on \emph{linear probing} evaluation, in which only a classifier is tuned during the downstream training while the pre-trained representations are kept frozen. It reflects how the representations obtained by the pre-trained models are linearly separable \wrt semantic categories. 

As shown in \cref{tbl:appdix-lp}, the \emph{linear probing} performance is consistently lower than the \emph{fine-tuning} performance. Coupled with the case that \emph{linear probing} does not save much training time for evaluating lightweight models, it is not a proper way to utilize the pre-trained models compared to the \emph{fine-tuning} setting. 

Furthermore, the \emph{linear probing} evaluation results do not reflect fine-tuned performance according to \cref{tbl:appdix-lp} and \cref{tbl:transfer}, especially for those downstream tasks with relatively sufficient labeled data, \eg, iNat18, ImageNet, thus may lead to an underestimation of the value of some pre-trained models in the practical utility on downstream tasks. We attribute it to that \emph{linear probing} only evaluates the final representation of the pre-trained models, which makes it overlook the value of providing good initialization for lower layers. For instance, MAE-Tiny is better at it than MoCov3-Tiny. 

Additionally, the inferior \emph{linear probing} results of MAE-Tiny to MoCov3-Tiny also support our analyses in \cref{sec:rep-analyses} that MoCov3-Tiny learns more semantics at an abstract level relevant to recognition in higher layers than MAE-Tiny. But our proposed distillation technique can improve the results to a certain extent.

\begin{table}[t]
\input{appdix/lp}
\vspace{-15pt}
\end{table}
\begin{table}[t]
\input{appdix/pretrain-more}
\vspace{-15pt}
\end{table}
\begin{table}[h!]
\input{appdix/transfer-more}
\vspace{-15pt}
\end{table}

\subsection{Analyses for More Self-Supervised Pre-Training Methods}\label{sec:appdix-more}

In the main paper, our analyses mainly focus on MAE \cite{mae} and MoCov3 \cite{mocov3}. In this section, more self-supervised pre-training methods are involved. Specifically, another MIM-based method, SimMIM \cite{simmim}, and another CL-based method, DINO \cite{dino}, are evaluated based on the lightweight ViT-Tiny. The 400-epoch pre-trained models are denoted as SimMIM-Tiny and DINO-Tiny respectively. 

We first evaluate their downstream performance on ImageNet and other classification tasks, and object detection and segmentation tasks, as shown in \cref{tbl:pretrain-more} and \cref{tbl:transfer-more}. They are also revised versions of \cref{tbl:pretrain} and \cref{tbl:transfer} in the main paper. According to the results, we find that MIM-based methods are generally superior to CL-based methods on data-sufficient tasks, \eg, ImageNet and iNat18, while inferior on data-insufficient tasks. Downstream data scale matters for all these methods and none of them achieve consistent superiority on all downstream tasks.

\begin{figure*}[t!]
    \centering
    \includegraphics[width=1.0\textwidth]{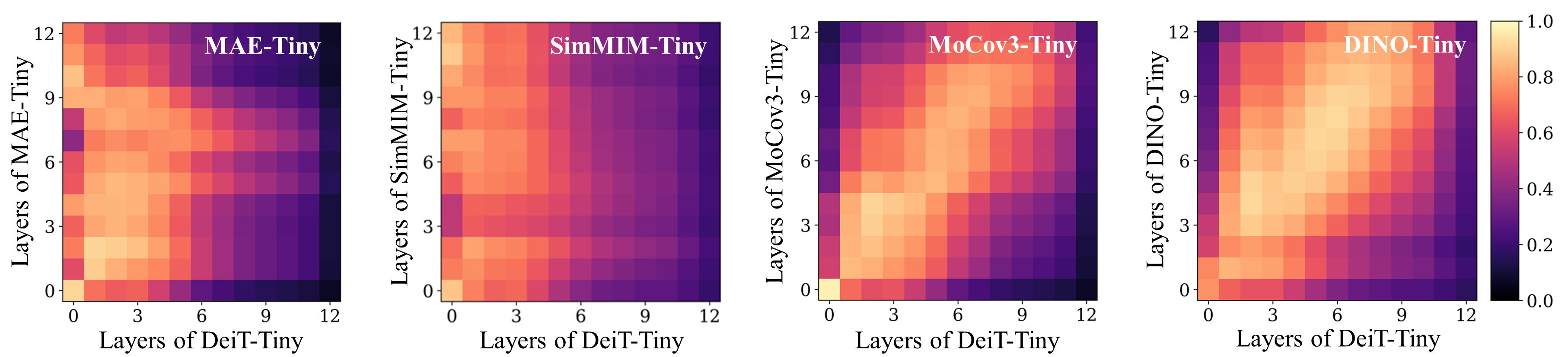}
    \vspace{-15pt}
    \caption{\small{\textbf{Layer representation analyses} for more self-supervised pre-trained models.}}
    \label{fig:appdix-cmp-more}
\vspace{-5pt}
\end{figure*}

Then we explore the layer representation of these models by CKA-based similarity analyses, as shown in \cref{fig:appdix-cmp-more}. We observe similar layer representation structures for both MIM family and CL family. For instance, SimMIM-Tiny also learns poor semantics on higher layers. 

\begin{figure*}[t!]
    \centering
    \includegraphics[width=1.0\textwidth]{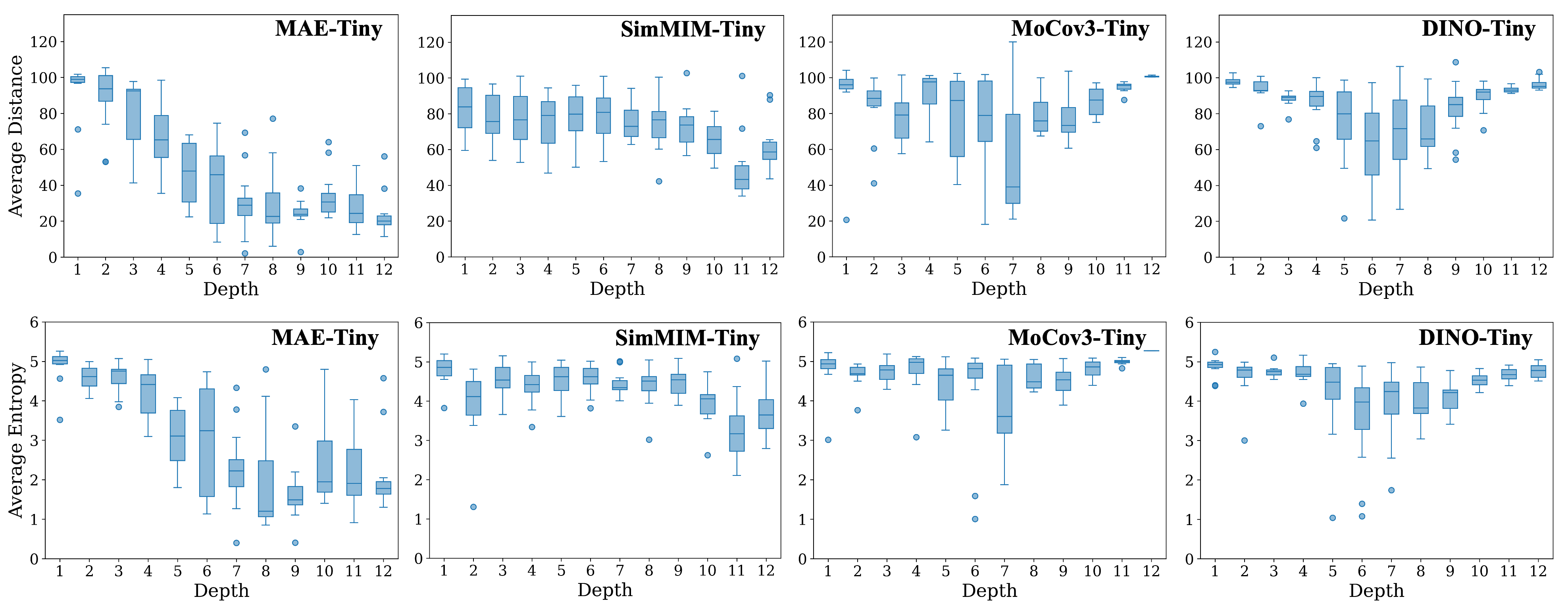}
    \vspace{-15pt}
    \caption{\small{\textbf{Attention analyses} for more self-supervised pre-trained models.}}
    \label{fig:appdix-attn-more}
\vspace{-5pt}
\end{figure*}

Finally, we carry out the attention analyses for these models, as shown in \cref{fig:appdix-attn-more}. We also observe consistent properties for MIM family and CL family. SimMIM-Tiny also tends to focus on local patterns with concentrated attention in higher layers like MAE-Tiny, while DINO-Tiny behaves like MoCov3-Tiny and has broad and global attention in higher layers.

\section{More Analyses on Distillation}


\begin{figure*}
    \centering
    \includegraphics[width=0.8\textwidth]{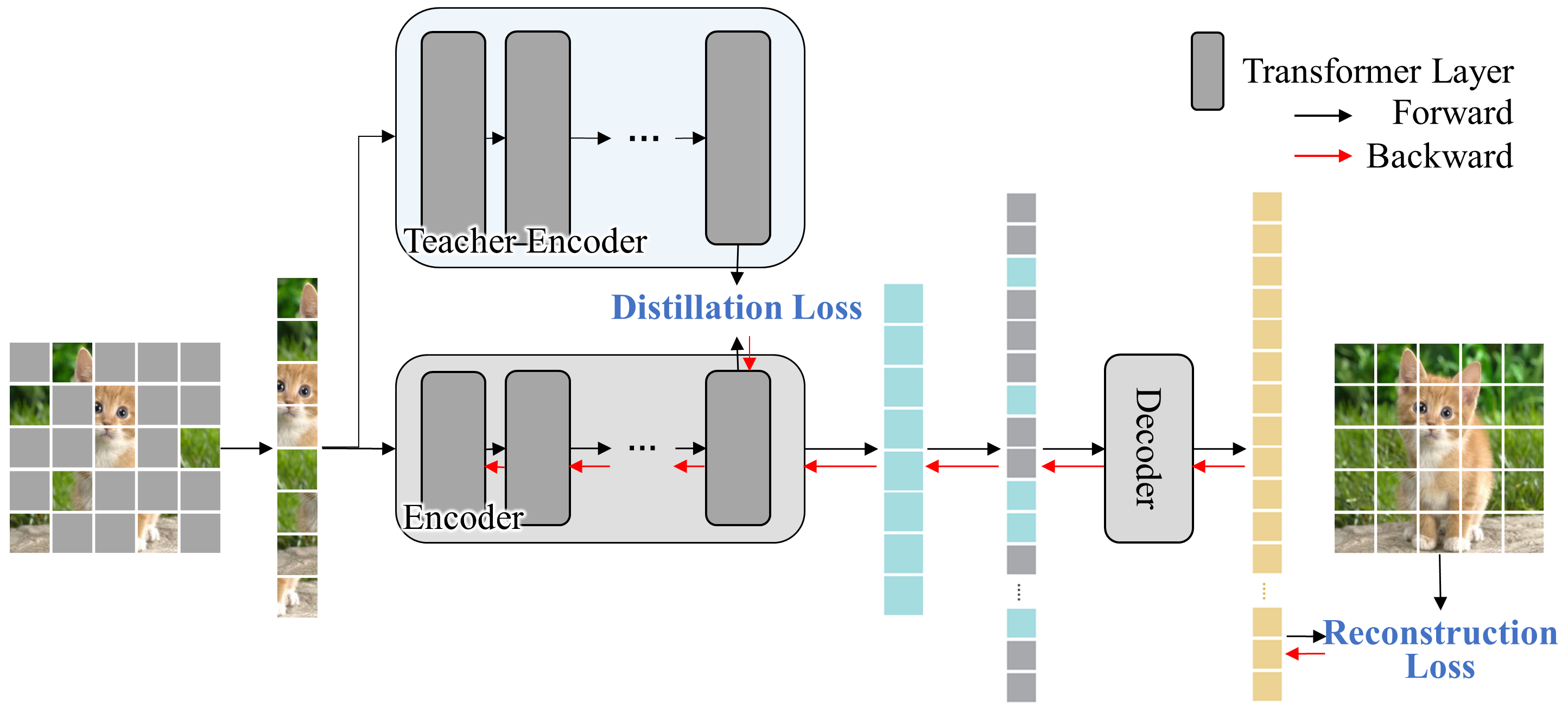}
    \vspace{-5pt}
    \caption{\small{\textbf{Illustration of the distillation process.}}}
    \label{fig:appdix-arch}
\end{figure*}
\subsection{Illustration of the Distillation Process}
We illustrate our distillation process in \cref{fig:appdix-arch} for a better presentation and explanation.

Based on the mask auto-encoder, we introduce a teacher ViT, which is pre-trained with MAE. During pre-training, the teacher processes the same visible image patches as the student encoder, and the attention-based distillation loss is calculated between the attention maps of the corresponding teacher's and student's layers. The parameters of the student are updated based on the joint backward gradients from the distillation loss and the original MAE's reconstruction loss, while the teacher's parameters remain frozen throughout the pre-training process.

\subsection{Attention Map Analyses for the Distilled Pre-trained Models}
\begin{figure*}[ht]
    \centering
    \includegraphics[width=0.80\textwidth]{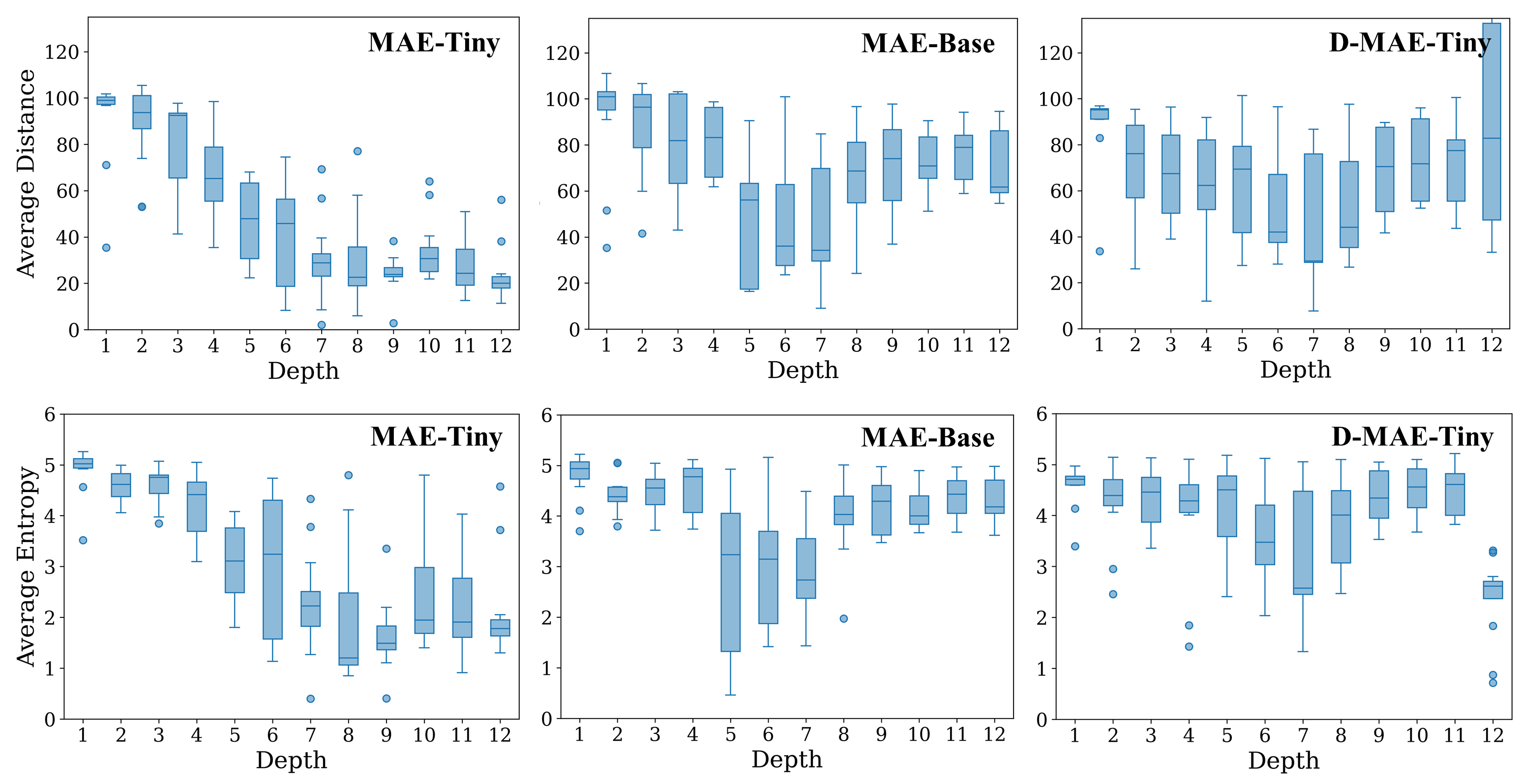}
    \vspace{-10pt}
    \caption{\small{\textbf{Attention distance and entropy analyses} for the distilled MAE-Tiny.}}
    \label{fig:appdix-attn-distill}
\vspace{-10pt}
\end{figure*}
we analyze the attention distance and entropy of the distilled MAE-Tiny introduced in \cref{sec:distill} (D-MAE-Tiny), which is only applied distillation on the attention map of the last layer during the pre-training with MAE. As shown in \cref{fig:appdix-attn-distill}, we observe more global and broad attention in the higher layers of D-MAE-Tiny compared with MAE-Tiny, which behaves more like the teacher, MAE-Base. We attribute it to that the distillation on the final layer (\ie, the 12th layer) forces the distilled layer of the student to imitate the teacher's attention and also requires the several preceding layers to make changes to meet the imitation. We reckon that it may be useful to capture semantic features and improve downstream performance. 

We also find the attention distance of the last layer shows more diversity: some attention heads are rather global and the others are local, and all of them are concentrated. We reckon that it shows odd behaviors for the reason that the layer can not handle both training targets from the reconstruction task and distillation restricted to the model size. But the more plentiful supervision indeed improves the quality of previous layers and thus achieves better downstream performance.

\subsection{Applying Distillation on More Networks}\label{sec:appdix-distill}

To further evaluate our proposed distillation method, we additionally apply it to the pre-training of ViT-Small also with MAE-Base as the teacher. The configurations of these models are presented in \cref{tbl:appdix-vit}. The transfer evaluation results are presented in \cref{tbl:appdix-small}. The transfer performance of the distilled MAE-Small (D-MAE-Small) surpasses the baseline model, MAE-Small by a large margin, which shows the efficacy of the distillation.

\subsection{Distilling with Larger Teachers}

\begin{figure*}[t]
    \centering
    \includegraphics[width=0.80\textwidth]{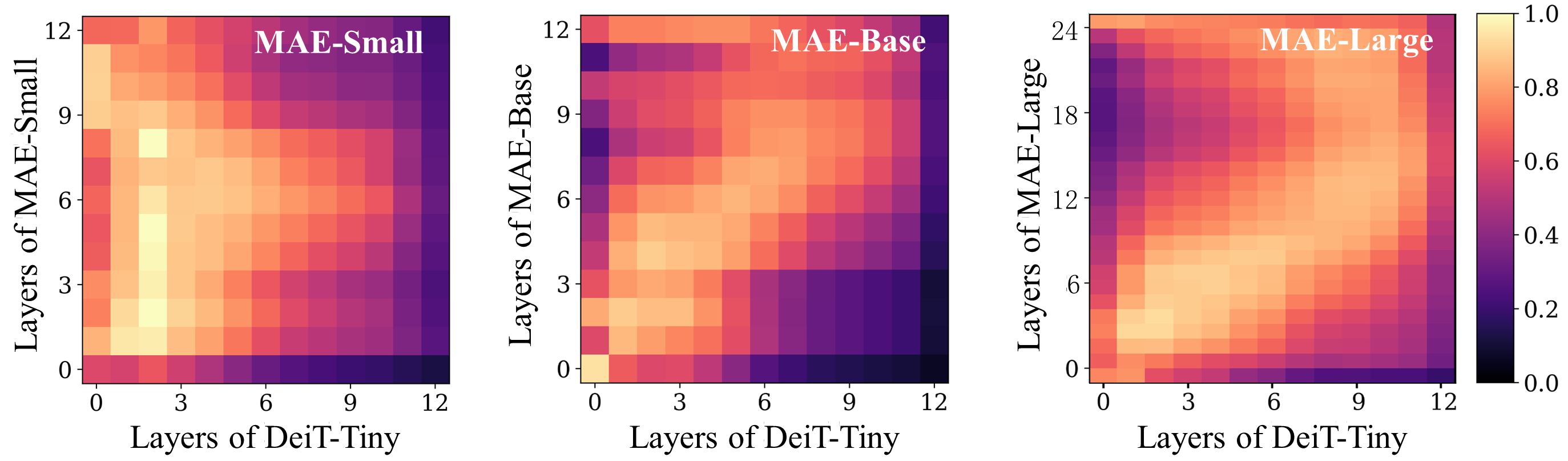}
    \vspace{-5pt}
    \caption{\small{\textbf{Layer representation analyses} of the teachers for distillation.}}
    \label{fig:appdix-cmp-tch}
\vspace{-10pt}
\end{figure*}

We further conduct additional experiments with various models as teachers and compared their performance on various downstream tasks (see \cref{tbl:appdix-tch}). The configurations of the student model (ViT-Tiny) and teacher models are presented in \cref{tbl:appdix-vit}. The results indicate that \emph{an appropriately sized teacher model provides the most improvement gains in distillation}, which is a common finding in the area of knowledge distillation \citep{cho2019efficacy,jin2019knowledge,mirzadeh2020improved}. To further investigate the impact of teacher size, we conducted CKA-based layer representation analyses of these teachers, as shown in \cref{fig:appdix-cmp-tch}. It can be seen that a teacher that is too small (MAE-Small) also suffers from degraded representation on higher layers and can not provide sufficient knowledge, while a teacher that is too large (MAE-Large) would result in a mismatch of capacity with the tiny student model, considering it has over 50 times more parameters than ViT-Tiny with different depths and attention head numbers, which leads to a little distinct learned pattern compared to the reference tiny model, and may not be suitable for the student. 

\begin{table}[th]
\input{appdix/vit}
\end{table}
\begin{table}[th]
\input{appdix/small}
\vspace{-5pt}
\end{table}
\begin{table}[t]
\input{appdix/larger-tch}
\end{table}

%% file: appdix/eval-detail.tex
\setlength{\tabcolsep}{4pt}
\begin{center}
\renewcommand{\arraystretch}{1.0} 
{
\caption{\small{\textbf{Fine-tuning evaluation settings}.}}
\label{tbl:appdix-eval-detail}
\small
\begin{tabular}{c|c}
\toprule
\textbf{config} & \textbf{value}\\
\midrule
optimizer & AdamW\\
base learning rate & 1e-3 \\
weight decay & 0.05 \\
optimizer momentum & $\beta_1, \beta_2=0.9, 0.999$ \\
layer-wise \emph{lr} decay \citep{beit} & 0.85 (MAE), 0.75 (MoCo-v3) \\
batch size & 1024 \\
learning rate schedule & cosine decay \citep{sgdr} \\
warmup epochs & 5 \\
training epochs & \{100, 300, 1000\} \\
augmentation & RandAug(10, 0.5) \citep{randaug}  \\
colorjitter & 0.3 \\
label smoothing & 0 \\
mixup \citep{mixup} & 0.2 \\
cutmix \citep{cutmix} & 0 \\
drop path \citep{huang2016deep} & 0 \\
\bottomrule
\end{tabular}}
\end{center}

%% file: appdix/mocov3.tex
\setlength{\tabcolsep}{4pt}
\begin{center}
\renewcommand{\arraystretch}{1.0} 
{
\caption{\small{\textbf{Pre-training setting for MoCo-v3}.}}
\label{tbl:appdix-mocov3}
\small
\begin{tabular}{c|c}
\toprule
\textbf{config} & \textbf{value}\\
\midrule
optimizer & AdamW \\
base learning rate & 1.5e-4 \\
weight decay & 0.1 \\
optimizer momentum & $\beta_1, \beta_2=0.9, 0.999$\\
batch size & 1024 \\
learning rate schedule & cosine decay \\
warmup epochs & 40 \\
training epochs & 400 \\
momentum coefficient & 0.99 \\
temperature & 0.2\\
\bottomrule
\end{tabular}}
\end{center}

%% file: appdix/transfer-detail.tex
\setlength{\tabcolsep}{4pt}
\begin{center}
\renewcommand{\arraystretch}{1.0} 
{
\caption{\small{\textbf{Transfer evaluation details}.}}
\label{tbl:appdix-transfer-detail}
\small
\begin{tabular}{c|c|c|c}
\toprule
\textbf{Dataset} & \textbf{Learning rate} & \textbf{Total epochs and warm-up epochs} & \textbf{layer-wise \emph{lr} decay} \\
\midrule
Flowers & \{0.01, 0.03, 0.1\} & \{(150,30),(250,50)\} & \{1.0, 0.75\} \\
Pets & \{0.01, 0.03, 0.1\} & \{(70,14),(150,30)\} & \{1.0, 0.75\}\\
Aircraft & \{0.01, 0.03, 0.1\} & \{(50,10),(100,20)\} & \{1.0, 0.75\}\\
Cars & \{0.01, 0.03, 0.1\} & \{(50,10),(100,20)\} & \{1.0, 0.75\}\\
CIFAR100 & \{0.03, 0.1, 0.3\} & \{(25, 5),(50,10)\} & \{1.0, 0.75\}\\
\bottomrule

\end{tabular}}
\vspace{-15pt}
\end{center}

%% file: appdix/lp.tex
\begin{center}

\caption{\small{\textbf{Linear probing evaluation} of pre-trained models on downstream classification tasks. Top-1 accuracy is reported.
}}
\label{tbl:appdix-lp}
\small
\begin{tabular}{c|ccccccc}
\toprule
\diagbox{\small\textbf{Init.}}{\small\textbf{Datasets}} & \textbf{Flowers} & \textbf{Pets} & \textbf{Aircraft} & \textbf{Cars} & \textbf{CIFAR100} & \textbf{iNat18} & \textbf{ImageNet} \\
 \midrule
 \colorgray{\textit{supervised}} \\
DeiT-Tiny & 91.0 & \textbf{92.0} & 41.2 & \textbf{47.9} & \textbf{73.6} & \textbf{39.8} & - \\
 \midrule
\colorgray{\textit{self-supervised}} \\
MoCov3-Tiny & \textbf{93.2} & 83.5 & \textbf{44.8} & 44.5 & 73.4 & 36.2 & \textbf{62.1} \\
MAE-Tiny & 48.9 & 25.0 & 12.8 & 8.8 & 31.0 & 1.4 & 23.3 \\
\rowcolor{lightgray}
D-MAE-Tiny & 77.1 & 55.5 & 20.1 & 16.4 & 58.4 & 10.7 & 42.0 \\
\toprule
\end{tabular}
\end{center}

%% file: appdix/pretrain-more.tex
\begin{center}
\renewcommand{\arraystretch}{1.0} 
{
\caption{\small{\textbf{Comparisons on more pre-training methods}. It is a revised version of \cref{tbl:pretrain} in the main paper with more self-supervised pre-training methods.
}
}
\label{tbl:pretrain-more}
\small
\begin{tabular}{cccc|cc}
\toprule
\multicolumn{4}{c|}{\textbf{\ \quad\quad Pre-training}} & \multicolumn{2}{c}{\textbf{\textbf{Fine-tuning}}}\\
\textbf{Methods} & \textbf{Data} & \textbf{Epochs} & \textbf{Time (hour)} & \textbf{recipe} & \textbf{Top-1 Acc. (\%)} \\
\midrule
from scratch & - & - & - & ori. & 74.5 \\
from scratch & - & - & - & impr. & 75.8 \\
Supervised \citep{steiner2021train} & IN21K w/ labels & 30 & 20 & impr. & 76.9 \\
Supervised \citep{steiner2021train} & IN21K w/ labels & 300 & 200 & impr. & 77.8 \\
MoCo-v3 \citep{mocov3} & IN1K w/o labels & 400 & 52 & impr. & 76.8 \\
MAE \citep{mae} & IN1K w/o labels & 400 & 23 & impr. & 78.0 \\
DINO \cite{dino} & IN1K w/o labels & 400 & 83 & impr. & 77.2 \\
SimMIM \cite{simmim} & IN1K w/o labels & 400 & 40 & impr. & 77.9 \\
\rowcolor{lightgray}
D-MAE-Tiny (ours) & IN1K w/o labels & 400 & 26 & impr. & \textbf{78.4} \\
\toprule
\end{tabular}}
\end{center}

%% file: appdix/transfer-more.tex
\setlength{\tabcolsep}{4pt}
\begin{center}
\renewcommand{\arraystretch}{0.9} 
{
\caption{\small{\textbf{Transfer evaluation on classification tasks and dense-prediction tasks for more pre-training methods}. It is a revised version of \cref{tbl:transfer} in the main paper with more self-supervised pre-training methods.
}}
\label{tbl:transfer-more}
\small
\begin{tabular}{c|cccccccc}
\toprule
\multirow{2}{*}{\diagbox{\textbf{Init.}}{\textbf{Datasets}}} & \textbf{Flowers} & \textbf{Pets} & \textbf{Aircraft} & \textbf{Cars} & \textbf{CIFAR100} & \textbf{iNat18} & \textbf{COCO}\small{(det.)} & \textbf{COCO}\small{(seg.)}\\
& \scriptsize{(2k/6k/102)} & \scriptsize{(4k/4k/37)} & \scriptsize{(7k/3k/100)} & \scriptsize{(8k/8k/196)} & \scriptsize{(50k/10k/100)} & \scriptsize{(438k/24k/8142)} & \multicolumn{2}{c}{\scriptsize{(118k/50k/80)}}\\
 \midrule
\colorgray{\textit{supervised}} \\
DeiT-Tiny & \textbf{96.4} & \textbf{93.1} & 73.5 & 85.6 & \textbf{85.8} & 63.6 & 40.4 & 35.5 \\
\midrule
\colorgray{\textit{self-supervised}} \\
MoCov3-Tiny & 94.8 & 87.8 & 73.7 & 83.9 & 83.9 & 54.5 & 39.7 & 35.1 \\
MAE-Tiny & 85.8 & 76.5 & 64.6 & 78.8 & 78.9 & 60.6 & 39.9 & 35.4 \\
DINO-Tiny & 95.6 & 89.3 & 73.6 & 84.5 & 84.7 & 58.7 & 41.4 & 36.7 \\
SimMIM-Tiny & 77.2 & 68.9 & 55.9 & 70.4 & 77.7 & 60.8 & 39.3 & 34.8 \\
\rowcolor{lightgray}
D-MAE-Tiny (ours) & 95.2 & 89.1 & \textbf{79.2} & \textbf{87.5} & 85.0 & \textbf{63.6} & \textbf{42.3} & \textbf{37.4} \\
\toprule
\end{tabular}}
\end{center}

%% file: appdix/vit.tex
\setlength{\tabcolsep}{12pt}
\begin{center}

\caption{\small{
\textbf{Configurations of ViTs}.
}}
\label{tbl:appdix-vit}
\begin{threeparttable}
\small
\begin{tabular}{c|ccccccc}
\toprule
\textbf{Model} & \textbf{channel dimension} & \textbf{\#heads} & \textbf{\#layers} & \textbf{\#params} \\
\midrule
ViT-Tiny & 192 & 12 & 12 & 6M \\
ViT-Small & 384 & 12\tnote{\ddag} & 12 & 22M \\
ViT-Base & 768 & 12 & 12 & 86M \\
ViT-Large & 1024 & 16 & 24 & 304M \\
\toprule
\end{tabular}
\begin{tablenotes}
    \footnotesize
    \item[\ddag] Our ViT-Small is with heads=12 following \citet{mocov3}.
    \end{tablenotes}
\end{threeparttable}
\end{center}

%% file: appdix/small.tex
\begin{center}

\caption{\small{
\textbf{Distillation on MAE-Small}. Top-1 accuracy for the transfer performance on downstream classification tasks of pre-trained models w. or w/o. distillation is reported.
}}
\label{tbl:appdix-small}
\small
\begin{tabular}{c|ccccccc}
\toprule
\diagbox{\small\textbf{Init.}}{\small\textbf{Datasets}} & \textbf{Flowers} & \textbf{Pets} & \textbf{Aircraft} & \textbf{Cars} & \textbf{CIFAR100} & \textbf{iNat18} & \textbf{ImageNet} \\
 \midrule
 \colorgray{\textit{supervised}} \\
DeiT-Small & \textbf{97.4} & \textbf{94.2} & 77.6 & 88.2 & \textbf{89.2} & 66.5 & 80.2 \\
 \midrule
\colorgray{\textit{self-supervised}} \\
MAE-Small & 91.2 & 82.0 & 65.8 & 79.2 & 80.8 & 63.2 & 82.1 \\
\rowcolor{lightgray}
D-MAE-Small & 95.8 \tiny\colorred{(+4.6)} & 91.4 \tiny\colorred{(+9.4)} & \textbf{80.7} \tiny\colorred{(+14.9)} & \textbf{88.3} \tiny\colorred{(+9.1)} & 87.8 \tiny\colorred{(+7.0)} & \textbf{66.9} \tiny\colorred{(+3.7)} & \textbf{82.5} \tiny\colorred{(+0.4)} \\
\toprule
\end{tabular}
\end{center}

%% file: appdix/larger-tch.tex
\begin{center}

\caption{\small{\textbf{Distillation with different sized teachers}. Top-1 accuracy for the transfer performance on downstream classification tasks of the distilled pre-trained models is reported.
}}
\label{tbl:appdix-tch}
\small
\begin{tabular}{cc|ccccccc}
\toprule
\multicolumn{2}{c|}{\textbf{Pre-training}} & \multicolumn{7}{c}{\textbf{Fine-tuning}} \\
\textbf{Student} & \textbf{Teacher} & \textbf{Flowers} & \textbf{Pets} & \textbf{Aircraft} & \textbf{Cars} & \textbf{CIFAR100} & \textbf{iNat18} & \textbf{ImageNet} \\
 \midrule
 MAE-Tiny & - & 85.8 & 76.5 & 64.6 & 78.8 & 78.9 & 60.6 & 78.0 \\
 MAE-Tiny & MAE-Small & 89.4 & 78.6 & 65.2 & 78.9 & 79.6  & 61.5 & 78.1 \\
MAE-Tiny & MAE-Base & \textbf{95.2} & \textbf{89.1} & \textbf{79.2} & \textbf{87.5} & \textbf{85.0} & \textbf{63.6} & \textbf{78.4} \\
MAE-Tiny & MAE-Large & 94.0 & 87.3 & 77.1 & 85.2 & 84.2 & 63.1 & 78.3 \\
\toprule
\end{tabular}
\end{center}